\PassOptionsToPackage{table}{xcolor}
\documentclass{article}

\newcommand{\paperversion}{preprint}

\usepackage{fancyhdr}
\usepackage{natbib}
\usepackage{dependency/iclr2027_conference}
\usepackage{etoolbox}
\ifdefstring{\paperversion}{anonymous}{}{%
    \ifdefstring{\paperversion}{camera-ready}{\iclrfinalcopy}{%
        \ifdefstring{\paperversion}{preprint}{\iclrfinalcopy}{%
            \PackageError{VEI}{Unknown paper version '\paperversion'}%
                {Use anonymous, camera-ready, or preprint.}%
        }%
    }%
}
\usepackage{times}

\usepackage{latexsym}
\usepackage[T1]{fontenc}
\usepackage[utf8]{inputenc}
\usepackage{microtype}
\usepackage{inconsolata}
\usepackage{graphicx}

\usepackage{amsmath,amsfonts,bm,mathrsfs,amsthm}
\usepackage{algorithm}
\usepackage{algorithmic} %
\usepackage{amssymb}
\usepackage{wasysym}
\usepackage{wrapfig,lipsum,booktabs}

\def\eqref#1{equation~\ref{#1}}

\def\1{\bm{1}}

\DeclareMathAlphabet{\mathsfit}{\encodingdefault}{\sfdefault}{m}{sl}
\SetMathAlphabet{\mathsfit}{bold}{\encodingdefault}{\sfdefault}{bx}{n}

\usepackage{enumitem}
\usepackage{multirow}
\usepackage{makecell}
\usepackage{xcolor}
\usepackage{caption}

\usepackage[most]{tcolorbox}
\newtcolorbox{promptbox}[1]{
    enhanced,
    width=\textwidth,
    colback=white,
    colframe=black,
    boxrule=0.8pt,
    arc=2pt,
    left=6pt,
    right=6pt,
    top=5pt,
    bottom=5pt,
    fontupper=\ttfamily\small,
    title=\textbf{#1},
    coltitle=white,
    colbacktitle=black,
    colframe=black,
    attach boxed title to top left={xshift=0pt,yshift=0pt},
    boxed title style={
        sharp corners,
        boxrule=0pt,
        colback=black,
        left=6pt,
        right=6pt,
        top=2pt,
        bottom=2pt
    }
}

\usepackage{hyperref}
\usepackage{url}
\usepackage{cleveref}

\title{Video Evidence Indexing: Learning Where to Look from Video Previews for Token-Budgeted Long-Video Question Answering}

\author{Haowen Guan, Shengzhi Li, Shichao Pei}

\begin{document}
\maketitle
\ifdefstring{\paperversion}{preprint}{%
    \fancyhead{}%
    \renewcommand{\headrulewidth}{0pt}%
}{}

\begin{abstract}

Long-video question answering is limited by the high cost of visual tokens and by the fixed context width of current VLMs. A long-video question may require broad temporal coverage, but the answer is often supported by only a compact set of moments. To locate these moments efficiently, we propose token-budgeted Video Evidence Indexing (VEI): given a dense low-resolution Video Preview, the model constructs a compact high-resolution Evidence Set for final reasoning. We treat VEI as a policy that must jointly solve \textit{evidence localization}, which finds question-relevant moments, and \textit{budget planning}, which decides where to spend the limited high-resolution frame budget. We implement this idea with an inference pipeline: the Video Preview provides cheap global coverage, Video Evidence Indexing constructs the Evidence Set, and Answer Generation combines both inputs for final VQA. To address missing frame-level supervision, we adopt privileged self-distillation, where an answer-aware teacher guides the normal test-time policy on student-generated indexing traces. We explore previews at 1, 6, 12, and 24 visual tokens per frame, training a single policy that supports all four resolutions. Experiments show that Video Evidence Indexing improves accuracy under limited visual budgets, and self-distillation further improves both QA accuracy and temporal evidence localization.
\end{abstract}

\section{Introduction}
\label{sec:introduction}

Recent advances in Video Language Models (VLMs) \citep{Qwen3.5,Qwen3-vl,Gemini2.5} have shown strong capabilities in understanding and reasoning over complex video content. However, long-video question answering (VQA) is still largely handled by fitting as much video input as possible into a finite context width. A long video may contain thousands of frames, while the answer to a question is often supported by only a compact set of moments \citep{VUSurvey2023,LVBench2024}. Processing the full video at high resolution provides broad temporal coverage and fine visual detail, but the visual-token cost quickly becomes too large. Uniformly sampling fewer frames reduces this cost, but it may miss the short event, object state, or scene change that determines the answer.

\begin{figure*}[!t]
    \vspace{-2px}
    \centering
    \includegraphics[width=\textwidth]{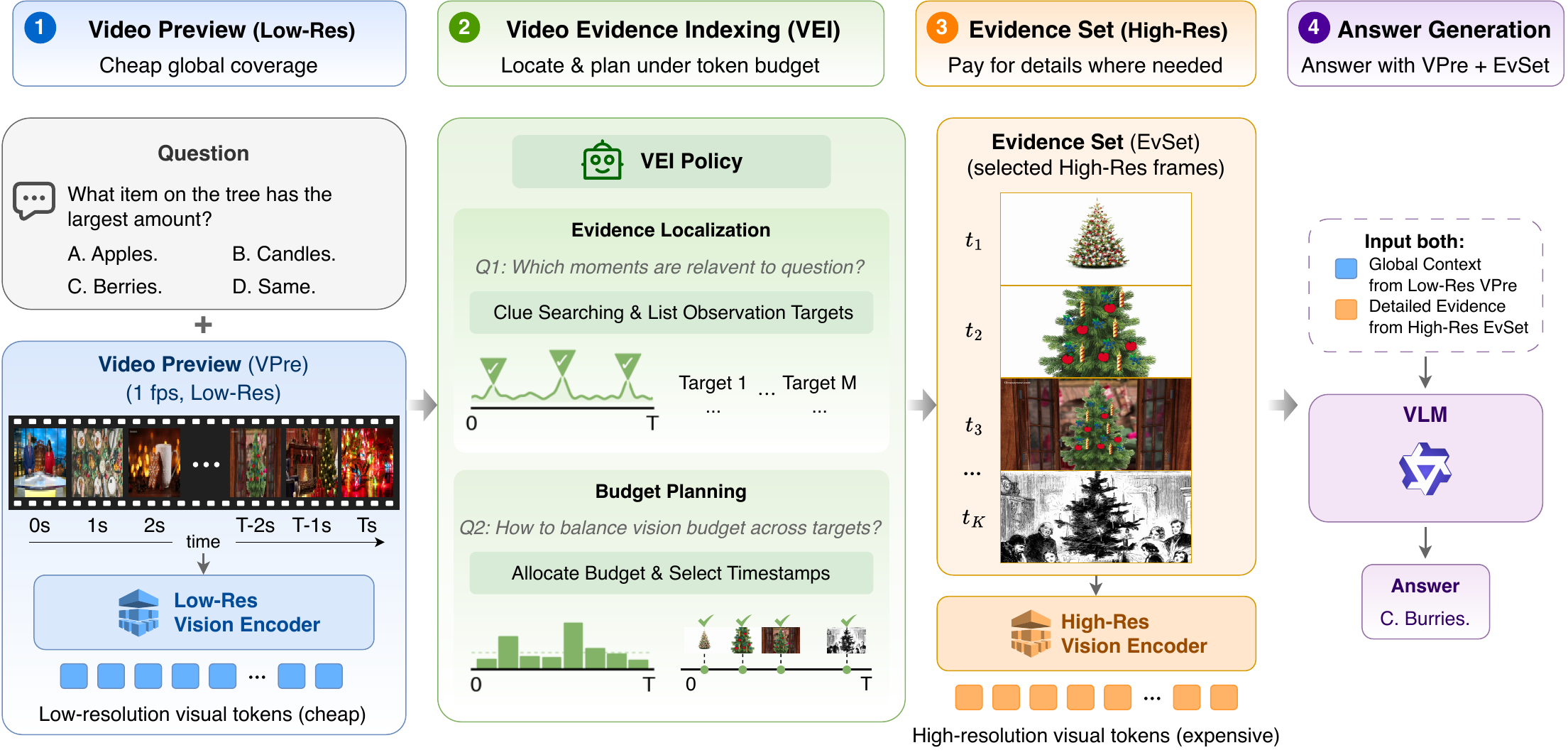}
    \caption{\textbf{Overview of Video Evidence Indexing.}
    A densely sampled low-resolution Video Preview (VPre) provides cheap global coverage under a strict context limit. We explore how VEI can use VPre to locate complementary evidence and reserve high-resolution detail for the moments that matter.}
    \label{fig:method_overview}
    \vspace{-12px}
\end{figure*}

Token budget matters not only because fewer visual tokens reduce inference cost, but also because context width is a hard capacity limit. On common 24--48GB GPUs, an 8B VLM can often handle only tens of thousands of tokens before memory and latency become limiting. Long-video QA therefore needs methods that operate within a fixed context width, not only methods that reduce average cost. Token-compression methods reduce visual redundancy through pruning, merging, or compression inside the model \citep{TokenCompressionSurvey2026,DToMA2025,DyTok2025}, but they often require architectural changes or access to internal token states. Adaptive frame selection uses relevance, coverage, attention, or inference feedback to construct compact evidence sets~\citep{AKS2025,MEC2026,DAFS2026,GridProbe2026}. These signals offer efficient selection, while leaving open how to train explicit evidence requests for the downstream question. Agentic video-observation methods plan and refine evidence acquisition through repeated or parallel tool calls~\citep{LensWalk2026,ReViSe2026,ParaVT2026}. Their observation structure differs from our single preview-to-index phase.

\begin{wrapfigure}{r}{0.48\textwidth}
    \centering
    \includegraphics[width=\linewidth]{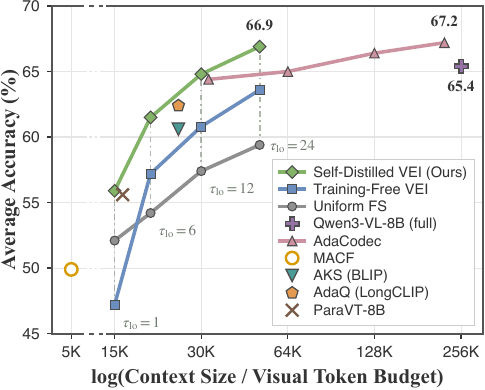}
    \vspace{-9px}
    \captionsetup{font=small}
    \caption{\textbf{Accuracy--context trade-off.}
    Mean accuracy across three datasets (Section~\ref{sec:experiments}). All methods use Qwen3-VL-8B. Comparators cover predictive compression (AdaCodec), multi-agent aggregation (MACF), encoder-based selection (AKS, AdaQ), and RL-trained video-tool acquisition (ParaVT).}
    \label{fig:accuracy_context}
    \vspace{-12px}
\end{wrapfigure}

We study how best to use a global video preview to guide the acquisition of question-relevant detail under a limited context budget. Specifically, we explore densely sampled \textbf{Video Previews} (VPre) at four spatial resolutions $\tau_{\text{lo}}=1, 6, 12,$ and $24$ visual tokens per frame (Figure~\ref{fig:method_overview}, \ref{fig:accuracy_context}). Even when fine details are lost, low-resolution frames often preserve coarse cues such as scene layout, object presence, motion, and event boundaries. These cues may be insufficient for final answer verification, but they can help the model decide where to spend high-resolution tokens. We formulate this setting as token-budgeted \textbf{Video Evidence Indexing} (VEI). Given a densely sampled low-resolution VPre, the model performs VEI by mapping the preview and question to a compact \textbf{Evidence Set} (EvSet) under a fixed token budget. The goal is not to find a single salient keyframe, but to construct a question-relevant set of high-resolution frames that supports final reasoning while balancing coverage, diversity, and redundancy reduction.

Our formulation turns VQA into a policy-learning problem. A pretrained VLM has general visual reasoning ability, but it is not specifically trained to allocate a limited visual budget from a blurry preview. A strong VEI policy must jointly solve \textit{evidence localization}, which identifies question-relevant moments, and \textit{budget planning}, which decides where limited high-resolution frames should be spent. However, most VQA datasets provide answer labels without corresponding frame-level evidence annotations, making direct supervised fine-tuning (SFT)~\citep{SFT2023} difficult. Reinforcement learning with answer-level rewards, such as GRPO~\citep{GRPO2024}, can optimize final-answer correctness, but provides little direct guidance for individual selection decisions. A correct answer alone does not validate the preceding evidence choices, so even successful rollouts cannot reliably serve as reference labels for evidence localization and budget planning.

To guide these intermediate decisions without frame-level labels, we adopt privileged self-distillation~\citep{OPSD2026}. A frozen teacher receives the video preview and correct answer, and is instructed to identify visible evidence supporting that answer. This provides pseudo-supervision for evidence localization and budget planning: the teacher can generate reference indexing traces for the student to learn from through off-policy distillation. In practice, we let the student generate its own indexing traces while the teacher supplies token-level guidance along them, following prior work~\citep{OPD2024}. This on-policy setting, also favored by our ablations, aligns supervision with the decisions the student actually encounters, teaching it what evidence to seek and how to distribute its limited observations. Our contributions are summarized as follows:
\begin{itemize}[leftmargin=8pt]
    \item We explore the use of dense Video Previews for long-video question answering across four preview token budgets, demonstrating their potential to guide evidence acquisition under limited context.

    \item We formulate \textbf{Video Evidence Indexing (VEI)} as a policy that couples evidence localization with budget planning, using explicit observation targets and target-wise timestamp selection to acquire complementary evidence.

    \item We address missing frame-level labels by deriving supervision for VEI decisions from an answer-aware teacher. Privileged self-distillation transfers this guidance across preview resolutions.
\end{itemize}

\section{Related Work}
\label{sec:related_work}

\paragraph{Efficient video representations and memory.}
Token pruning and merging reduce visual redundancy~\citep{TokenCompressionSurvey2026,DToMA2025,DyTok2025,UnifiedCompression2026,WaveZip2026}, while learned compressors and reusable memories retain compact video context~\citep{Tempo2026,QViCMF2026,MemoryCard2026,CFD2026}. AdaCodec encodes detailed reference frames and compact inter-frame changes~\citep{AdaCodec2026}; MACF distributes segment perception across agents communicating through latent messages~\citep{MACF2026}. These approaches compress representations or aggregate observations. VEI instead generates explicit evidence requests from a visual preview and reloads selected frames, preserving global coverage while reserving detailed visual tokens for question-relevant evidence.

\paragraph{Frame selection and budget allocation.}
Encoder-based selectors use question relevance, coverage, or diversity~\citep{AKS2025,QFrame2025,AdaQ2026,MEC2026,MarKey2026}; MEC constructs a ranking reusable across budgets. Other approaches use confidence-guided exploration~\citep{FOCUS2025}, timestamped textual grounding~\citep{GCR2026}, or learned contextual selectors~\citep{FFS2025,RLFrameSelection2026}. DAFS extracts frame scores from VLM attention and allocates candidate and resolution budgets~\citep{DAFS2026}, while EcoFrame and GridProbe use inference feedback or multi-frame confidence probes~\citep{EcoFrame2026,GridProbe2026}. ReUnit allocates mixed-granularity visual layouts~\citep{ReUnit2026}. VEI uses a dense visual preview to formulate explicit observation targets and distribute timestamps across them. This lets the policy allocate its limited frames according to what each target requires, with the aim of covering the different evidence needed to answer the question.

\paragraph{Reasoning-driven video acquisition.}
Agentic methods reason about which evidence to inspect through temporal tools and adaptive observation~\citep{VITAL2025,VideoMTR2025,FrameThinker2025,ZoomZero2025,VideoZoomer2025,LongVT2025,VideoMem2025,AVP2025}. LensWalk plans temporal scope and sampling density, while ReViSe maintains a compact state across rounds~\citep{LensWalk2026,ReViSe2026}. A.I.R. and VideoXAgent further adapt evidence acquisition to the question~\citep{AIR2026,VideoXAgent2026}. ParaVT learns parallel video-tool calls with reinforcement learning~\citep{ParaVT2026}; AutoSkill routes questions to discovered selection programs~\citep{AutoSkill2026}. VEI plans all evidence requests from one dense preview before answering, avoiding repeated search rounds while keeping the high-resolution frame budget fixed.

\paragraph{Privileged visual self-distillation.}
On-policy distillation supervises student-generated prefixes~\citep{OPD2024}; self-distillation can derive guidance from richer teacher context~\citep{OPSD2026,OPCD2026,SDPO2026}. Video-OPSD and Clue-OPSD use annotated evidence frames or intervals to strengthen video supervision~\citep{VideoOPSD2026,ClueOPSD2026}. RP-OPSD, S$^2$VOPD, and Vision-OPD exploit resolution, augmentation, or regional-view asymmetries~\citep{RPOPSD2026,S2VOPD2026,VisionOPD2026}. These methods primarily distill perception and answer reasoning from visual inputs. VEI distills the evidence-acquisition policy, teaching the model what to inspect and how to distribute high-resolution budget. This helps the student use a compressed preview to acquire question-relevant detail under a limited context, without frame-level annotations.

\section{Method}
\label{sec:method}

We study multiple-choice long-video QA under a fixed visual-token budget \(B\). Given a video \(V\), a question \(Q\), and answer choices \(\mathcal{A}=\{a_1,\ldots,a_M\}\), the model must acquire the evidence needed to answer within this budget. As shown in Figure~\ref{fig:method_overview}, our inference pipeline follows four steps: Video Preview, Video Evidence Indexing, Evidence Set construction, and Answer Generation. The VEI policy performs two coupled behaviors: \textit{evidence localization} identifies what visual evidence to look for, and \textit{budget planning} decides where to spend the limited high-resolution frames. We then train these behaviors through privileged self-distillation.

\begin{figure*}[!t]
    \vspace{-5px}
    \centering
    \includegraphics[width=\textwidth]{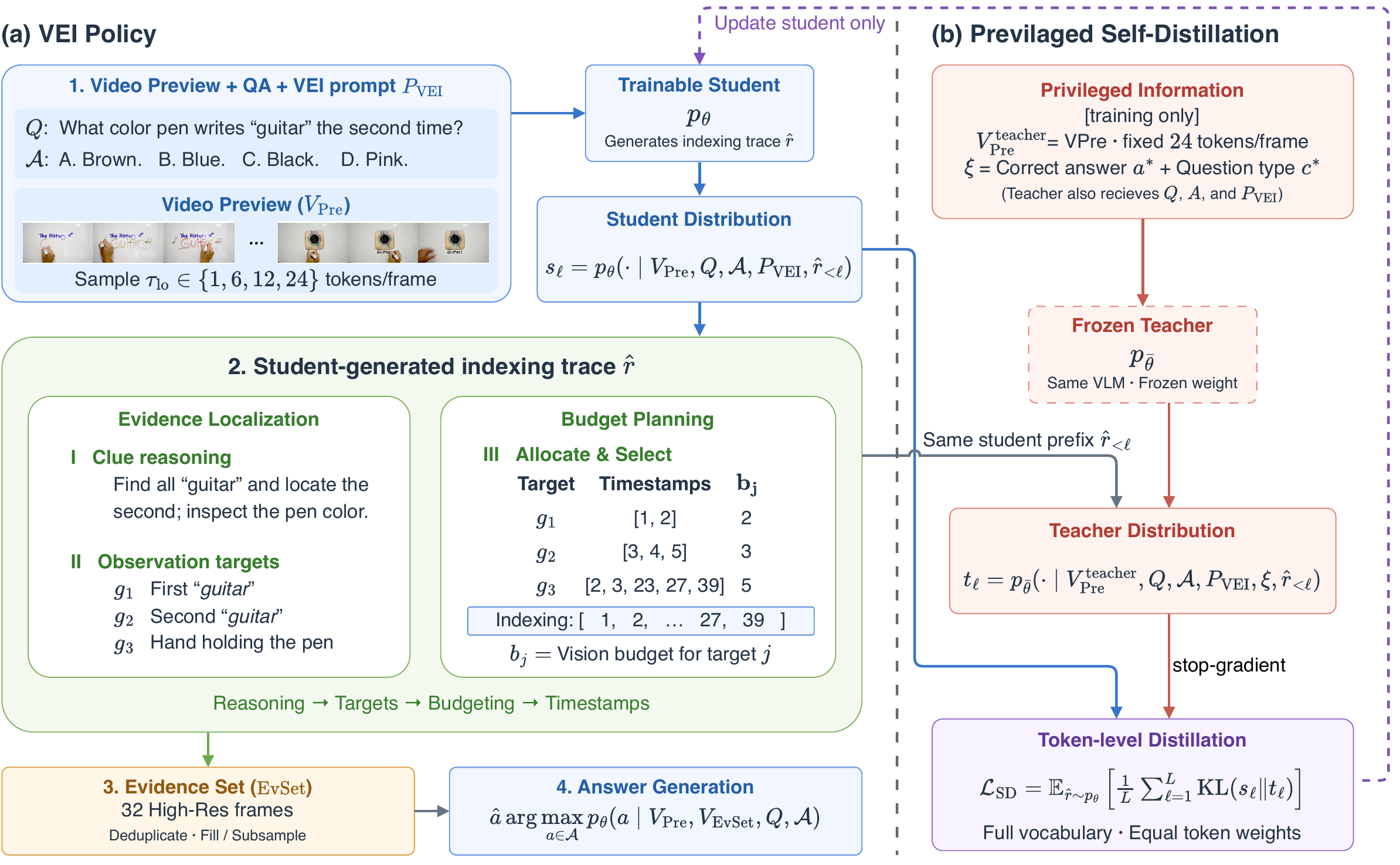}
    \caption{\textbf{Privileged self-distillation for VEI.}
    (a) The student generates observation targets and timestamp lists; the complete VEI prompt $P_{\text{VEI}}$ appears in \cref{fig:selection_prompt}. (b) Privileged information enables the teacher to guide the student's search toward evidence supporting the correct answer. Distilling this guidance teaches the student what to look for and how to allocate its frame budget from compressed previews, without frame-level annotations or privileged inputs at inference.}
    \label{fig:privileged_self_distillation}
    \vspace{-8px}
\end{figure*}

\subsection{Video Preview}
\label{sec:vision_compression}

We construct the Video Preview by sampling the video at \(1\) fps and reducing each frame's spatial resolution. Let \(t_1,\ldots,t_n\) be these sampling timestamps in seconds, and let \(V(t_i)\) denote the original frame at \(t_i\). The preview is
\begin{equation}
V_{\mathrm{Pre}}
=\bigl(\operatorname{Downsize}_{\tau_{\mathrm{lo}}}(V(t_i))\bigr)_{i=1}^{n},
\end{equation}
where \(\operatorname{Downsize}_{\tau_{\mathrm{lo}}}\) reduces each frame's resolution so that the backbone's visual encoder produces at most \(\tau_{\mathrm{lo}}\) visual tokens. With \(n\) sampled frames, the preview uses at most \(n\tau_{\mathrm{lo}}\) visual tokens. Sampling is reduced only if required by the context limit.

The temporally dense preview provides global context at low visual-token cost. It is designed to preserve coarse event progression across nearby frames, supporting global reasoning and localization of complementary high-resolution evidence in one indexing attempt.

\subsection{Video Evidence Indexing}
\label{sec:coarse_to_fine}

The Video Evidence Indexing Prompt \(P_{\mathrm{VEI}}\) guides the model to identify evidence needs and distribute its high-resolution observations across them (Appendix~\ref{sec:appendix_prompt}). Given this prompt, the Video Preview, question, and answer choices, the policy generates an indexing trace:
\begin{equation}
\hat r\sim p_{\theta}(\cdot\mid V_{\mathrm{Pre}},Q,\mathcal A,P_{\mathrm{VEI}}).
\end{equation}
The trace contains two coupled behaviors: evidence localization and budget planning.

\paragraph{Evidence localization.}
The model briefly reasons about what to inspect from the question and preview, then generates a list of observation targets:
\begin{equation}
G=[g_1,\ldots,g_J]\sim p_{\theta}(\cdot\mid V_{\mathrm{Pre}},Q,\mathcal A,P_{\mathrm{VEI}}).
\end{equation}
Each \(g_j\) describes a specific evidence need in natural language, such as an object's appearance, an action, or an event to distinguish. The prompt requests compact, diverse targets and encourages merging overlapping ones. This explicit list organizes what must be verified and guides the subsequent allocation of frames.

\paragraph{Budget planning.}
Conditioned on the target list, the model jointly plans its high-resolution observations through target-specific timestamp lists:
\begin{equation}
(E_1,\ldots,E_J)\sim p_{\theta}(\cdot\mid V_{\mathrm{Pre}},Q,\mathcal A,P_{\mathrm{VEI}},G),
\qquad b_j=|E_j|.
\end{equation}
The visual budget \(b_j\) for target \(g_j\) is implicit in its timestamp choices and list termination. Joint planning aims to balance target coverage and reduce redundancy within \(K\) frames, reserving more observations for targets requiring temporal comparison or fine detail.

\paragraph{Evidence Set and answer generation.}
We enforce the final \(K\)-frame budget by merging valid timestamps into \(E=\bigcup_{j=1}^{J}E_j\), removing duplicates, and uniformly supplementing or subsampling. We reload these frames at high resolution to form \(V_{\mathrm{EvSet}}\). A separate call generates the answer from the preview and Evidence Set:
\begin{equation}
\hat a=\arg\max_{a\in\mathcal A}
p_{\theta}(a\mid V_{\mathrm{Pre}},V_{\mathrm{EvSet}},Q,\mathcal A).
\end{equation}
With a per-frame visual-token cap of \(\tau_{\mathrm{hi}}\) for the Evidence Set, the answering call's visual input costs at most \(n\tau_{\mathrm{lo}}+K\tau_{\mathrm{hi}}\) tokens. We choose the preview and evidence budgets so that \(n\tau_{\mathrm{lo}}+K\tau_{\mathrm{hi}}\le B\).

\subsection{Privileged Self-Distillation}
\label{sec:opsd}

The lack of frame-level evidence annotations in most VQA datasets makes direct supervision of intermediate VEI decisions difficult. To improve evidence localization and budget planning without explicit annotations, we adopt privileged self-distillation~\citep{OPSD2026}, using answer-aware token-level guidance to train the VEI policy (Figure~\ref{fig:privileged_self_distillation}).

\paragraph{Student indexing traces.}
For each example, we uniformly sample the student's preview budget \(\tau_{\mathrm{lo}}\) from \(\{1,6,12,24\}\) visual tokens per frame. The student receives \(V_{\mathrm{Pre}},Q,\mathcal A,P_{\mathrm{VEI}}\) and generates an \(L\)-token indexing trace with clue reasoning, observation targets, and timestamp lists:
\begin{equation}
\hat{r}=[z_1,\ldots,z_L]
\sim p_{\theta}(\cdot\mid V_{\mathrm{Pre}},Q,\mathcal A,P_{\mathrm{VEI}}).
\end{equation}
Here, \(z_{\ell}\) is the token at position \(\ell\). This trains one VEI policy for all four preview resolutions. Our rollout-source ablation favors student-generated (on-policy) traces over teacher-generated (off-policy) traces among the tested settings (Section~\ref{sec:self_distillation_analysis}).

\paragraph{Privileged teacher.}
The teacher starts from the same backbone as the student and remains frozen. It receives a preview of the same video fixed at \(24\) tokens per frame, denoted \(V_{\mathrm{Pre}}^{\mathrm{teacher}}\), along with the question, choices, and VEI prompt. Its privileged information \(\xi=(a^{\star},c^{\star})\) contains the correct answer \(a^{\star}\) and optional complementary information \(c^{\star}\), such as the question type. The teacher uses this context to score the student's existing trace rather than generate a separate reference. At each token position \(\ell\), their next-token distributions are
\begin{equation}
\begin{aligned}
s_{\ell}&=p_{\theta}(\cdot\mid V_{\mathrm{Pre}},Q,\mathcal A,P_{\mathrm{VEI}},\hat r_{<\ell}),\\
t_{\ell}&=p_{\bar\theta}(\cdot\mid V_{\mathrm{Pre}}^{\mathrm{teacher}},Q,\mathcal A,P_{\mathrm{VEI}},\xi,\hat r_{<\ell}),
\end{aligned}
\end{equation}
where \(\bar\theta\) denotes the frozen teacher parameters and \(\hat r_{<\ell}=[z_1,\ldots,z_{\ell-1}]\) is the shared prefix. We train the student distribution \(s_{\ell}\) toward the teacher distribution \(t_{\ell}\) at each prefix. This provides answer-aware self-correction: the student first seeks evidence, then learns from the teacher's guidance on its own decisions, including those following a weak target or an unhelpful timestamp.

\paragraph{Distillation objective.}
We minimize reverse KL between student and teacher distributions over the indexing trace, excluding final-answer generation:
\begin{equation}
\mathcal{L}_{\mathrm{SD}}
=
\mathbb{E}_{\hat{r}\sim p_{\theta}(\cdot\mid V_{\mathrm{Pre}},Q,\mathcal A,P_{\mathrm{VEI}})}
\left[
\frac{1}{L}\sum_{\ell=1}^{L}
\mathrm{KL}(s_{\ell}\|t_{\ell})
\right].
\end{equation}
The divergence is computed over the full vocabulary and averaged equally across token positions, without special weights for timestamps. Gradients update only the student; sampled tokens and teacher distributions are held fixed during each update. Among the tested objectives, our ablation favors reverse KL over forward KL and generalized Jensen--Shannon divergence (Section~\ref{sec:self_distillation_analysis}).

The same objective provides guidance for both VEI behaviors. At reasoning and target positions, it guides how the student describes the evidence needed by the question. At timestamp positions, it guides where to look. Guidance over list endings is central to budget planning: whether to request another frame or close the list determines how many frames are proposed for that target. Together, these signals are designed to transfer evidence localization and budget allocation without per-target quota labels or a separate allocation loss. At inference, the student follows the same VEI pipeline with no privileged information.

\section{Experiments}
\label{sec:experiments}

\suppressfloats[t]
\begin{table*}[t]
\vspace{-1pt}
\centering
\resizebox{\textwidth}{!}{
\begin{tabular}{l|cc|cccccc|cc}
\toprule
\multirow{2}{*}{Method}
& \multirow{2}{*}{\makecell{Context\\Limit}}
& \multirow{2}{*}{\makecell{Preview Size\\$(\#\mathrm{Frame} / \mathrm{Tok})$}}
& \multicolumn{4}{c}{VideoMME {\scriptsize w/o sub.}}
& LVBench
& MLVU{\tiny MCQ}
& \multicolumn{2}{c}{Average} \\
\cline{4-9} \cline{10-11}
& &
& \scriptsize Short
& \scriptsize Medium
& \scriptsize Long
& \scriptsize Overall
& \scriptsize Train
& \scriptsize Test
& \scriptsize Score
& \scriptsize \% \\
\hline

\rowcolor{gray!13}
\multicolumn{11}{c}{\textit{Use 100\% Context Width Across All Benchmarks}} \\
Qwen3-VL-8B
& $256\mathrm{K}$
& $2048/640_{\mathrm{Tok}}$
& $77.8$ & $70.7$ & $64.7$ & $71.0$ & $57.6$ & $67.7$
& $65.4$
& $100.0$ \\
\hline

\rowcolor{gray!13}
\multicolumn{11}{c}{\textit{Use 5.9\% Context Width Across All Benchmarks}} \\
\makecell{Uniform FS}
& $15\mathrm{K}$
& $1 \mathrm{fps}/1_{\mathrm{Tok}}$
& $73.1$ & $60.0$ & $58.4$ & $63.8$ & $40.2$ & $52.2$
& $52.1$
& $79.6$ \\

Training-Free VEI
& $15\mathrm{K}$
& $1 \mathrm{fps}/1_{\mathrm{Tok}}$
& $70.4$ & $53.9$ & $51.4$ & $58.6$ & $36.7$ & $46.2$
& $47.2$
& $72.1 \;{\color{blue}(-7.5)}$ \\

Self-Distilled VEI
& $15\mathrm{K}$ & $1 \mathrm{fps}/1_{\mathrm{Tok}}$
& $75.8$ & $61.8$ & $54.2$ & $63.9$ & $44.9$ & $58.8$ & $55.9$ & $85.4 \;{\color{red}(+5.8)}$ \\
\hline

\rowcolor{gray!13}
\multicolumn{11}{c}{\textit{Use 7.8\% Context Width Across All Benchmarks}} \\
\makecell{Uniform FS}
& $20\mathrm{K}$
& $1 \mathrm{fps}/6_{\mathrm{Tok}}$
& $76.2$ & $62.7$ & $57.2$ & $65.4$ & $42.1$ & $55.2$
& $54.2$
& $82.9$ \\

Training-Free VEI
& $20\mathrm{K}$
& $1 \mathrm{fps}/6_{\mathrm{Tok}}$
& $73.2$ & $64.1$ & $55.9$ & $64.4$ & $46.1$ & $61.2$
& $57.2$
& $87.5 \;{\color{red}(+4.6)}$ \\

Self-Distilled VEI
& $20\mathrm{K}$ & $1 \mathrm{fps}/6_{\mathrm{Tok}}$
& $77.0$ & $65.8$ & $56.2$ & $66.3$ & $49.6$ & $68.5$ & $61.5$ & $94.0 \;{\color{red}(+11.1)}$ \\
\hline

\rowcolor{gray!13}
\multicolumn{11}{c}{\textit{Use 11.7\% Context Width Across All Benchmarks}} \\
\makecell{Uniform FS}
& $30\mathrm{K}$
& $1 \mathrm{fps}/12_{\mathrm{Tok}}$
& $75.6$ & $66.4$ & $58.9$ & $67.0$ & $46.3$ & $59.0$
& $57.4$
& $87.7$ \\

Training-Free VEI
& $30\mathrm{K}$
& $1 \mathrm{fps}/12_{\mathrm{Tok}}$
& $75.6$ & $67.2$ & $59.0$ & $67.3$ & $51.6$ & $63.5$
& $60.8$
& $92.9 \;{\color{red}(+5.2)}$ \\

Self-Distilled VEI
& $30\mathrm{K}$ & $1 \mathrm{fps}/12_{\mathrm{Tok}}$
& $77.8$ & $67.2$ & $59.9$ & $68.3$ & $53.7$ & $72.5$ & $64.8$ & $99.1 \;{\color{red}(+11.4)}$ \\
\hline

\rowcolor{gray!13}
\multicolumn{11}{c}{\textit{Use 19.5\% Context Width Across All Benchmarks}} \\
\makecell{Uniform FS}
& $50\mathrm{K}$
& $1 \mathrm{fps}/24_{\mathrm{Tok}}$
& $75.6$ & $65.9$ & $60.6$ & $67.3$ & $50.7$ & $60.2$
& $59.4$
& $90.8$ \\

Training-Free VEI
& $50\mathrm{K}$
& $1 \mathrm{fps}/24_{\mathrm{Tok}}$
& $77.7$ & $70.3$ & $63.6$ & $70.5$ & $53.5$ & $66.7$
& $63.6$
& $97.1 \;{\color{red}(+6.3)}$ \\

Self-Distilled VEI
& $50\mathrm{K}$ & $1 \mathrm{fps}/24_{\mathrm{Tok}}$
& $78.9$ & $70.5$ & $63.8$ & $71.1$ & $57.3$ & $72.3$ & $66.9$ & $102.2 \;{\color{red}(+11.4)}$ \\

\bottomrule
\end{tabular}}
\vspace{-3px}
\caption{\textbf{Main results across preview budgets.}
One self-distilled policy supports all four preview resolutions, outperforming uniform sampling even at one token per frame under the same 32-frame EvSet budget. The \(1\) fps preview rate is reduced when the input exceeds the context limit.}
\label{tab:main_result}
\vspace{-13pt}
\end{table*}

\subsection{Implementation Details}
\label{sec:implementation}

\paragraph{Backbone and variants.}
We use Qwen3-VL-8B~\citep{Qwen3-vl} as the backbone model. \textbf{Training-Free VEI} denotes Training-Free Video Evidence Indexing, which uses the VEI inference pipeline without parameter updates. \textbf{Self-Distilled VEI} denotes Self-Distilled Video Evidence Indexing, which trains the preview-based indexing policy with privileged self-distillation and uses the trained checkpoint for both indexing and answering at test time.

\paragraph{Datasets.}
We evaluate on VideoMME, LVBench, and MLVU \citep{VideoMME2024,LVBench2024,MLVU2024}. VideoMME contains videos from roughly one minute to one hour. LVBench focuses on long videos and provides temporal evidence windows for localization analysis. For MLVU, we evaluate on the 502-question multiple-choice test set. We report percentage accuracy.

\paragraph{Inference protocol and settings.}
For low-budget settings, the model first samples the video at  \(1\) fps and compresses each preview frame to a fixed visual-token cap \(\tau_{\mathrm{lo}}\in\{1,6,12,24\}\). This dense low-resolution input serves as the Video Preview.
We reduce the preview sampling rate only when required by the context limit. For VEI methods, the model indexes question-relevant evidence timestamps from the Video Preview (VPre), and the selected Evidence Set (EvSet) is then re-injected at higher resolution for final answer prediction. The uniform baseline uses the same VPre and high-resolution frame budget, replacing EvSet with uniformly sampled frames. Unless otherwise stated, the additional EvSet budget is \(32\) frames with a maximum of \(128\) visual tokens per frame.

The full-context Qwen3-VL baseline uses at most $2048$ frames and $640$ tokens per frame, the default setting in the original report. The preview budgets \(\tau_{\mathrm{lo}}\in\{1,6,12,24\}\) are chosen to cover a wide range of compression levels while avoiding distorted aspect ratios from very small intermediate token grids. For example, \(2\) tokens can force a \(1{:}2\) or \(2{:}1\) grid, which distorts many videos. For evaluation, we use deterministic decoding with temperature \(0.0\), repetition penalty \(2.0\), thinking disabled, and a maximum generation length of \(1024\) tokens. Our evaluations use a single run per setting.

\paragraph{Self-distillation training.}
Self-Distilled VEI is trained on the MLVU training set. To fit training into available GPU memory, we keep videos shorter than 20 minutes, yielding 2044 multiple-choice questions and 495 short-answer questions. The frozen teacher uses 24 visual tokens per preview frame and receives the correct answer, supplemented by the question type when available. For each example, the student uses a preview budget sampled uniformly from \(\{1,6,12,24\}\) visual tokens per frame, without privileged information. We train for 1,500 steps with a batch size of 8 using full-vocabulary reverse-KL distillation. Training uses LoRA on four H100 GPUs and takes 40 hours. Hyperparameters appear in Table~\ref{tab:sd_training_config}; prompts are in Appendix~\ref{sec:appendix_prompt}.

\subsection{Main Results}
\label{sec:main_results}

Table~\ref{tab:main_result} shows that Training-Free VEI improves over uniform sampling on all three datasets at 12 and 24 preview tokens per frame, demonstrating that our VEI policy works out of the box on a pretrained VLM. However, the pretrained model struggles to select useful evidence from one-token previews.

Privileged self-distillation strengthens VEI under extreme preview compression, with substantial one-token gains over uniform sampling on LVBench and MLVU and comparable VideoMME overall accuracy. At 24 tokens per frame, Self-Distilled VEI reaches 102.2\% of the full-context baseline's average accuracy with less than a quarter of its context budget. Our privileged self-distillation training enables VEI to make effective use of highly compressed Video Previews.

\suppressfloats[t]
\begin{table}[t]
\vspace{-1pt}
\centering
\small
\setlength{\tabcolsep}{5pt}
\resizebox{\columnwidth}{!}{%
\begin{tabular}{lccccccc}
\toprule
\multirow{2}{*}{Model} & \multirow{2}{*}{Size} & \multirow{2}{*}{Token Budget} & \multicolumn{2}{c}{VideoMME} & \multirow{2}{*}{LVBench} & \multirow{2}{*}{MLVU} & \multirow{2}{*}{AVG} \\
\cmidrule(lr){4-5}
 & & & Long & Overall & & & \\
\hline
\rowcolor{gray!13}
\multicolumn{8}{c}{\textit{(a) Proprietary Models}} \\
GPT-4o & -- & $128\mathrm{K}$ & 65.3 & 71.9 & 64.4 & 64.6 & 67.0 \\
Gemini-2.5-Pro & -- & $1\mathrm{M}$ & -- & 87.0 & 69.2 & 81.2 & 79.1 \\
\hline
\rowcolor{gray!13}
\multicolumn{8}{c}{\textit{(b) Open-Source / Public Pretrained Models}} \\
LLaVA-OneVision \citep{LLaVA-OneVision} & 7B & $32\mathrm{K}$ & 46.7 & 58.2 & 26.9 & 50.5 & 45.2 \\
LongVA \citep{LongVA} & 7B & $128\mathrm{K}$ & 47.6 & 54.3 & 36.2 & 43.9 & 44.8 \\
InternVL2.5 \citep{InternVL2.5} & 8B & $64\mathrm{K}$ & -- & 66.9 & 38.4 & 53.7 & 53.0 \\
Qwen2.5-VL \citep{Qwen2.5-VL} & 7B & $128\mathrm{K}$ & 51.6 & 65.1 & 45.3 & 54.8 & 55.1 \\
NVILA \citep{NVILA} & 8B & $64\mathrm{K}$ & 54.8 & 64.2 & 44.0 & 54.7 & 54.3 \\
VideoLLaMA3 \citep{VideoLLaMA3} & 7B & $64\mathrm{K}$ & 54.1 & 66.2 & 45.3 & 56.9 & 56.1 \\
Qwen3-VL \citep{Qwen3-vl} & 8B & $256\mathrm{K}$ & \textbf{64.7} & 71.0 & \textbf{57.6} & 67.7 & 65.4 \\
\hline
\rowcolor{gray!13}
\multicolumn{8}{c}{\textit{(c) Evidence-Acquisition Methods}} \\
MACF \citep{MACF2026} & 8B & $5\mathrm{K}$ & -- & 60.4 & 40.2 & 49.2 & 49.9 \\
ParaVT-8B \citep{ParaVT2026} & 8B & $16\mathrm{K}$ & -- & 62.1 & 39.8 & 65.0 & 55.6 \\
AKS + BLIP \citep{AKS2025} & 8B & $25\mathrm{K}$ & 59.0 & 68.6 & 49.0 & 64.3 & 60.6 \\
AdaQ + LongCLIP \citep{AdaQ2026} & 8B & $25\mathrm{K}$ & 59.4 & 69.6 & 51.4 & 66.2 & 62.4 \\
AdaCodec \citep{AdaCodec2026} & 8B & $64\mathrm{K}$ & -- & 70.6 & 57.2 & 67.2 & 65.0 \\
\hline
\rowcolor{gray!13}
\multicolumn{8}{c}{\textit{(d) VEI Cross-Backbone Evaluation}} \\
\multicolumn{8}{c}{\textit{LLaVA-OneVision-2-8B} \citep{LLaVA-OneVision-2}} \\
Uniform FS & 8B & $50\mathrm{K}$ & 56.0 & 65.9 & 47.5 & 57.4 & 56.9 \\
Training-Free VEI & 8B & $50\mathrm{K}$ & 58.3 & 68.4 & 53.5 & 64.9 & 62.3 \\
Self-Distilled VEI & 8B & $50\mathrm{K}$ & 58.8 & 68.7 & 55.8 & 65.1 & 63.2 \\
\multicolumn{8}{c}{\textit{Penguin-VL-8B} \citep{Penguin-VL}} \\
Uniform FS & 8B & $50\mathrm{K}$ & 51.7 & 63.9 & 39.3 & 56.0 & 53.1 \\
Training-Free VEI & 8B & $50\mathrm{K}$ & 57.2 & 66.6 & 44.5 & 60.6 & 57.2 \\
Self-Distilled VEI & 8B & $50\mathrm{K}$ & 57.7 & 67.2 & 47.4 & 64.0 & 59.5 \\
\multicolumn{8}{c}{\textit{Qwen3-VL-8B} \citep{Qwen3-vl}} \\
Uniform FS & 8B & $50\mathrm{K}$ & 60.6 & 67.3 & 50.7 & 60.2 & 59.4 \\
CLIP$^{\dagger}$ FS & 8B & $50\mathrm{K}$ & 62.1 & 68.9 & 53.9 & 65.5 & 62.8 \\
Training-Free VEI & 8B & $50\mathrm{K}$ & 63.6 & 70.5 & 53.5 & 66.7 & 63.6 \\
\rowcolor{cyan!15}
\textbf{Self-Distilled VEI} & 8B & $50\mathrm{K}$ & 63.8 & \textbf{71.1} & 57.3 & \textbf{72.3} & \textbf{66.9} \\
\bottomrule
\end{tabular}}
\vspace{-3px}
\caption{\textbf{Comparison with existing VLMs and evidence-acquisition methods.}
Prior results in (a--c) use published evaluations and budgets; methods in (c) use Qwen3-VL-8B. CLIP$^{\dagger}$ is our retrieval baseline (Appendix~\ref{sec:limitations_ethics}). Bold marks the best across (b--d).}
\label{tab:additional_vlm_comparison}
\vspace{-12pt}
\end{table}

\begin{figure*}[!b]
    \vspace{-12pt}
    \centering
    \includegraphics[width=\textwidth]{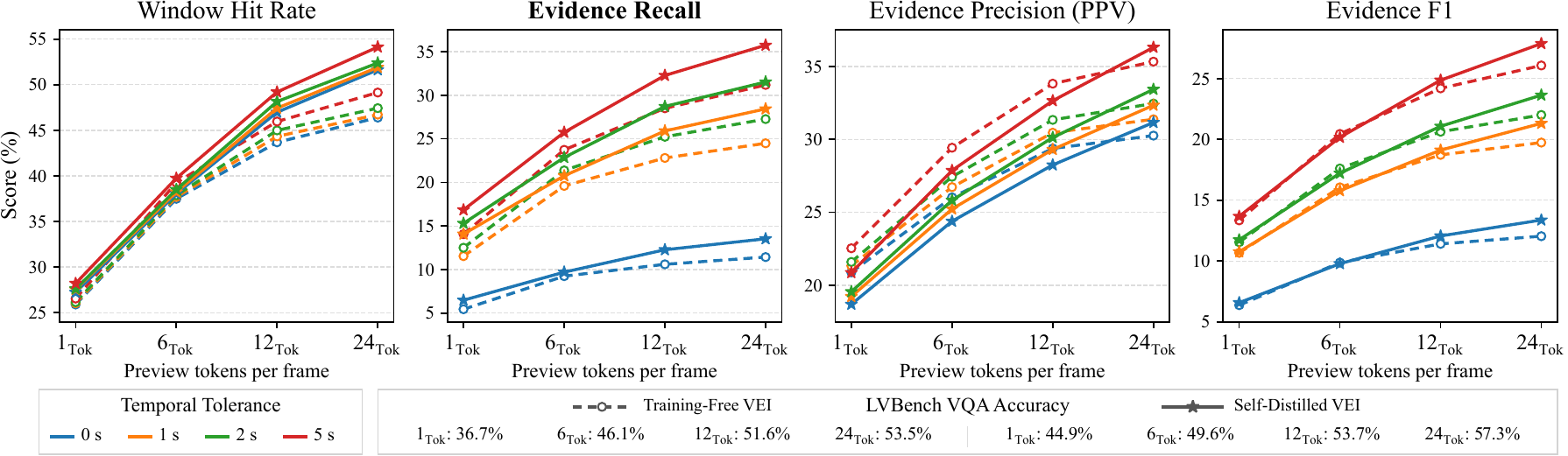}
    \vspace{-12pt}
    \caption{\textbf{Evidence localization on LVBench.}
    Training-Free and Self-Distilled VEI are compared across four preview budgets. Evidence Recall is the key metric, which measures how much of the annotated evidence the VEI policy recovers in its indexing.}
    \label{fig:lvbench_evidence_localization}
    \vspace{-0pt}
\end{figure*}

\vspace{-6pt}
\paragraph{Cross-backbone evaluation.}
\label{sec:cross_backbone}
Table~\ref{tab:additional_vlm_comparison}(d) shows that VEI's gains extend to LLaVA-OneVision-2 and Penguin-VL. Training-Free VEI improves all three benchmarks on both backbones, and Self-Distilled VEI further improves each result, supporting VEI beyond Qwen3-VL backbone.

\vspace{-3pt}
\noindent
\begin{minipage}[t]{0.49\textwidth}
\vspace{0pt}
\textbf{High-resolution evidence budget.}
We vary the high-resolution evidence budget over $K=8,16,32$ frames on VideoMME (Table~\ref{tab:evidence_budget}). Self-Distilled VEI remains robust under tighter budgets, surpassing 32-frame uniform sampling with only eight selected frames.
\end{minipage}\hfill
\begin{minipage}[t]{0.49\textwidth}
\vspace{0pt}
\centering
\captionsetup{font=small,justification=centering,singlelinecheck=false}
\footnotesize
\setlength{\tabcolsep}{3pt}
\begin{tabular*}{\linewidth}{@{\extracolsep{\fill}}lrrr@{}}
\toprule
Method & $K=8$ & $K=16$ & $K=32$ \\
\midrule
Uniform FS & 66.26 & 67.19 & 67.33 \\
Self-Distilled VEI & 69.52 & 70.24 & 71.07 \\
\bottomrule
\end{tabular*}
\captionof{table}{\textbf{Evidence-budget sensitivity on VideoMME.}}
\label{tab:evidence_budget}
\end{minipage}\par

\paragraph{Comparison with related works.}
\label{sec:efficiency_comparison}

Our CLIP$^{\dagger}$ baseline matches Training-Free VEI's preview and 32-frame high-resolution evidence budget (Table~\ref{tab:additional_vlm_comparison}(d)). VEI's higher average accuracy \mbox{($63.6\%$ vs.\ $62.8\%$)} supports structured indexing beyond relevance retrieval under matched visual budgets. Self-Distilled VEI reaches $66.9\%$ at $50\mathrm{K}$ context, exceeding AdaCodec's reported $65.0\%$ at $64\mathrm{K}$ visual tokens and approaching $67.2\%$ at $224\mathrm{K}$. Broader comparisons in (a--c) use different observation protocols; Uniform FS and CLIP$^{\dagger}$ provide matched visual-budget controls. AKS scores follow AdaQ's Qwen3-VL-8B evaluation~\citep{AdaQ2026}. Cost assumptions appear in Appendix~\ref{sec:appendix_cost}.

\subsection{Evidence Localization and Budget Planning}
\label{sec:component_ablation}

We separate the two VEI behaviors with a nested ablation. \textit{Base FS} directly selects timestamps from the question and preview. Adding evidence localization (EL) introduces explicit observation targets but retains one global timestamp list. Adding budget planning (BP) instead produces a timestamp list for each target. All variants use the same final evidence budget and uniform supplementation rule.

Table~\ref{tab:component_ablation} shows that listing targets alone brings only a modest gain without training and none after distillation. Target-wise selection adds 2.4 and 2.6 percentage points, respectively, while reducing uniform supplementation under the same final budget and filling rule. VEI remains strongest on questions with comparable selected-frame counts (Appendix~\ref{sec:appendix_comparable_counts}), supporting the value of connecting each target to concrete frame choices. Appendix~\ref{sec:appendix_evset_ablation} separately tests the contribution of high-resolution evidence by removing the Evidence Set while retaining the preview.

\subsection{Evidence Quality and Answer Correctness}
\label{sec:evset_quality}

Figure~\ref{fig:lvbench_evidence_localization} evaluates generated timestamps before filling or truncation against LVBench's evidence windows, expanded by one second per side and discretized into seconds, with \mbox{tolerance \(\delta\).} \mbox{\textbf{Window hit rate}} is the fraction of questions with a selection within \(\delta\) seconds of an evidence window. \textbf{Evidence recall} measures the fraction of evidence seconds covered within \(\delta\) of a selection; \textbf{evidence precision} measures the fraction of selections within \(\delta\) seconds of the evidence window. \textbf{Evidence F1} is their harmonic mean; these three metrics are averaged per question. Recall is central here because it measures how much supporting evidence is recovered. At one token per frame and \(\delta=5\) s, self-distillation raises recall from \(14.1\%\) to \(16.9\%\), supporting improved evidence coverage as an explanation for the QA gain from highly compressed previews.

\newsavebox{\veiAblationTop}
\newsavebox{\veiAblationBottom}
\newsavebox{\veiAblationRight}
\newlength{\veiAblationGap}
\setlength{\veiAblationGap}{12pt}

\begin{figure}[!t]
\centering
\captionsetup{font=small,justification=centering,singlelinecheck=false,skip=4pt}

\sbox{\veiAblationTop}{%
\begin{minipage}[b]{0.60\textwidth}
\centering
\footnotesize
\setlength{\tabcolsep}{1.5pt}
\begin{tabular*}{\linewidth}{@{\extracolsep{\fill}}lcccc@{}}
\toprule
& \multicolumn{2}{c}{Training-Free} & \multicolumn{2}{c}{Self-Distilled} \\
\cmidrule(lr){2-3}\cmidrule(lr){4-5}
Configuration & Accuracy & Filled frames & Accuracy & Filled frames \\
\midrule
Base FS & 63.2 & 17.6 & 69.7 & 17.5 \\
+ EL & 64.3 & 16.9 & 69.7 & 16.5 \\
+ EL + BP (VEI) & 66.7 & 10.7 & 72.3 & 9.8 \\
\bottomrule
\end{tabular*}
\captionof{table}{\textbf{VEI component ablation on MLVU.} Filled frames denotes the mean number of uniformly sampled frames added to reach the 32-frame evidence budget.}
\label{tab:component_ablation}
\end{minipage}}

\sbox{\veiAblationBottom}{%
\begin{minipage}[b]{0.60\textwidth}
\centering
\footnotesize
\setlength{\tabcolsep}{2pt}
\begin{tabular*}{\linewidth}{@{\extracolsep{\fill}}lcccc@{}}
\toprule
Method & \makecell{Hit\\rate} & \makecell{Evidence\\recall} & \makecell{Correct\\given hit} & \makecell{Correct\\given miss} \\
\midrule
Training-Free VEI & 48.8 & 31.2 & 65.6 & 41.6 \\
Self-Distilled VEI & 53.3 & 35.8 & 69.5 & 43.2 \\
Privileged teacher & 54.6 & 36.3 & 69.5 & 43.1 \\
\bottomrule
\end{tabular*}
\captionof{table}{\textbf{Evidence localization and QA on LVBench.} Scores (\%) use 24-token previews and a 5 s tolerance on 1,549 questions; recall measures generated timestamps as in Figure~\ref{fig:lvbench_evidence_localization}. The privileged teacher provides an empirical upper reference using answer information unavailable at inference.}
\label{tab:localization_qa}
\end{minipage}}

\sbox{\veiAblationRight}{%
\begin{minipage}[b]{0.38\textwidth}
\centering
\includegraphics[width=\linewidth]{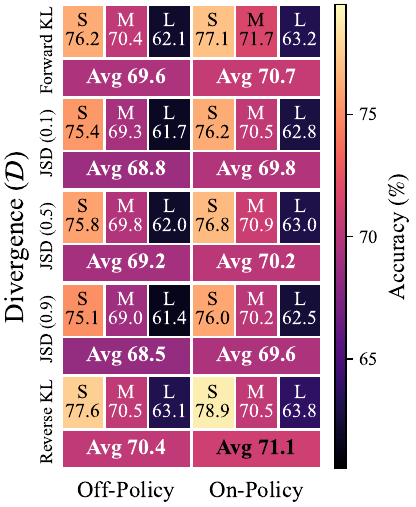}
\captionof{figure}{\textbf{Rollout policy and distillation objective ablation.} On-policy distillation with reverse KL achieves the highest overall VideoMME accuracy.}
\label{fig:hyperparameters}
\end{minipage}}

\begin{minipage}[t]{0.61\textwidth}
\vspace{0pt}
\usebox{\veiAblationTop}\par
\vspace{\veiAblationGap}
\nointerlineskip
\usebox{\veiAblationBottom}
\end{minipage}\hfill
\begin{minipage}[t]{0.38\textwidth}
\vspace{0pt}
\usebox{\veiAblationRight}
\end{minipage}
\end{figure}

Table~\ref{tab:localization_qa} uses the privileged teacher as an empirical upper reference for answer-aware localization. Self-Distilled VEI approaches the teacher's hit rate and evidence recall, and matches its answer accuracy on hit cases without privileged inputs.
Across all three policies, hits are associated with substantially higher answer accuracy, but do not guarantee correctness.

\subsection{Privileged Self-Distillation Analysis}
\label{sec:self_distillation_analysis}

\paragraph{Privileged guidance.}
The teacher diagnostic in Table~\ref{tab:localization_qa} reveals the evidence selections available under answer-aware conditioning; final answering remains non-privileged. During training, the teacher instead scores the student's prefixes as described in Section~\ref{sec:opsd}, transferring this guidance without supplying a separate gold trace or frame-level labels.

\paragraph{Rollout policy and objective.}
Figure~\ref{fig:hyperparameters} compares student-generated traces with teacher-generated traces. Student rollouts perform better in most settings, consistent with the benefit of supervising prefixes the student actually produces. This matters for VEI because later timestamps depend on earlier target choices. Reverse KL gives the strongest result in this comparison, followed by forward KL, while the JSD variants are slightly lower. We therefore use student rollouts and reverse KL for privileged self-distillation. The multi-resolution training setting additionally exposes one student to all four preview resolutions, while the teacher retains its 24-token preview.

\section{Conclusion}
\label{sec:conclusion}

We explore how dense, compressed Video Previews can guide evidence acquisition for long-video QA under limited context. VEI couples evidence localization with budget planning to construct a compact, question-relevant Evidence Set. We adopt privileged self-distillation to learn these decisions without frame-level annotations, producing one policy across four preview resolutions, including one token per frame. Our results demonstrate the value of using cheap global coverage to direct high-resolution observations toward the evidence needed for answering.

\clearpage

\subsection*{AI use statement}

The original manuscript was written by the authors. During revision, we used AI tools to edit figures, refine language and grammar. The authors reviewed and selectively incorporated AI-assisted suggestions, checking their consistency with the intended method and reported results. We take responsibility for the final text, claims, figures, and other content of this work.

\subsection*{Reproducibility statement}

Section~\ref{sec:method} describes the VEI pipeline and privileged self-distillation objective. Section~\ref{sec:implementation} specifies the datasets, preprocessing, inference settings, and training setup; Appendix~\ref{sec:appendix_prompt} provides the indexing prompt, and Table~\ref{tab:sd_training_config} lists training hyperparameters. We plan to publicly release the source code and the weights of our best-performing multi-resolution self-distilled model. These materials are being prepared and are not yet publicly available.

\bibliography{main}
\bibliographystyle{dependency/vei_references}

\clearpage

\appendix

\begin{center}
    {\LARGE Appendix}
\end{center}

\section{Component Ablation with Comparable Selection Counts}
\label{sec:appendix_comparable_counts}

The full-test-set ablation in Table~\ref{tab:component_ablation} gives every method 32 evidence frames, but VEI supplies more of these through its own selections and requires fewer uniform additions. To examine whether its advantage persists when this difference is smaller, we filter the same 502 MLVU questions by the number of frames each method selects.

\paragraph{Selection rule.}
For each question, we count each method's valid, unique selections before uniform filling and cap this count at 32. We retain the question if the capped counts for Base FS and + EL are each within $\pm3$ frames of + EL + BP (VEI). Thus, questions where all three independently select at least 32 frames automatically qualify. The original answering protocol is unchanged: excess selections are subsampled and shortages are uniformly supplemented to obtain exactly 32 evidence frames. We apply the filter separately to Training-Free and Self-Distilled models; within each row, all three configurations are evaluated on the same retained questions.

Before capping, all three methods select exactly the same number of frames on 67 Training-Free and 65 Self-Distilled questions. Separately, all three select at least 32 frames on 60 and 66 questions, respectively. Both conditions satisfy the inclusion rule above.

\begin{table}[!h]
\centering
\small
\begin{tabular*}{0.8\textwidth}{@{\extracolsep{\fill}}lcccc@{}}
\toprule
Selector & \makecell{Comparable\\questions} & Base FS & + EL & + EL + BP (VEI) \\
\midrule
Training-Free VEI & 220 / 502 & 70.0 & 66.4 & \textbf{74.6} \\
Self-Distilled VEI & 218 / 502 & 78.9 & 78.0 & \textbf{80.7} \\
\bottomrule
\end{tabular*}
\caption{\textbf{VEI component ablation with comparable selection counts.} Accuracy (\%) on shared questions within each row; every configuration uses a final 32-frame Evidence Set.}
\label{tab:comparable_count_ablation}
\end{table}

\paragraph{Targets need target-wise selection.}
Table~\ref{tab:comparable_count_ablation} shows that VEI outperforms both alternatives even when their selected-frame counts are similar. Base FS remains competitive, while adding targets without target-wise selection lowers accuracy on these subsets. This suggests that a target list alone does not ensure useful evidence acquisition: a single global list may fail to turn the planned targets into corresponding observations. Coupling localization with budget planning makes these targets actionable, recovering the advantage over direct selection in both models. The result supports this coupling, rather than treating target listing as an independently effective step.

\paragraph{Beyond the VEI format.}
Although self-distillation trains on VEI traces, the distilled model also performs well with Base FS and + EL. This is consistent with learned preview-use skills carrying over to simpler selection prompts. The Base FS scores of $70.0\%$ and $78.9\%$ are suggestive, although the different subsets prevent a paired estimate of the distillation gain.

\section{Prompt Details}
\label{sec:appendix_prompt}

This appendix records the first-turn Video Evidence Indexing Prompt. Final answer generation follows Evidence Set construction in a separate turn. During self-distillation, the privileged block conditions the teacher's scoring of student-generated prefixes; it is removed at test time.

\begin{figure*}[t]
\centering
\begin{promptbox}{Video-Evidence-Indexing Prompt}

\textbf{Input:}

\{Video Preview\}

Question: \{question\}

Possible answer choices:

\{options\}

\vspace{0.6em}
\textbf{[Optional] privileged teacher information:}

You are a teacher, and you are given the following information about the question.

The task type of this question is: \{question type\}.

The correct answer to this question is: \textbf{\{answer text\}}.

Please use this privileged information only to verify and refine a gold-standard
reference solution, not to shortcut the reasoning. Answer Steps 1--3
exactly as specified, with analysis that derives the correct choice
from visible evidence in the video. Finally, for each target, select a compact
and diverse set of evidence frames that supports this analysis, maximizes
question-relevant coverage, and minimizes redundancy, with at most \{maximum
frames per target\} frames per target.

\vspace{0.6em}
\textbf{Step 1: Evidence Planning}

Briefly answer: In order to locate the correct answer choice, what objects,
scenes, actions, or keyframes should we focus on?

\vspace{0.6em}
\textbf{Step 2: Observation Targets}

Based on Step 1, list a compact and diverse set of observation targets.
Merge overlapping targets when appropriate and avoid redundancy.

Do not give more than \{maximum number of targets\} targets.

\begin{itemize}
    \item Target 1: short description.
    \item Target 2: short description.
    \item Target 3: short description.
    \item \(\cdots\)
\end{itemize}

\vspace{0.6em}
\textbf{Step 3: Evidence Indexing}

For each observation target, index at most \{maximum frames per target\}
evidence frames from the Video Preview.

Output each indexed set as a Python list of absolute timestamps in seconds.

Example format:

\begin{itemize}
    \item Evidence Set 1: [12, 18, 23]
    \item Evidence Set 2: [45]
    \item Evidence Set 3: [61, 64]
    \item \(\cdots\)
\end{itemize}

Rules:
\begin{itemize}
    \item Each indexed set must be a Python list.
    \item Each entry must be an integer timestamp in seconds.
    \item Each list must contain at most \{maximum frames per target\} values.
    \item The indexed frames should be compact, diverse, and relevant to the target.
    \item Avoid repeated timestamps or near-duplicate frames unless they are needed for temporal verification.
\end{itemize}

\end{promptbox}
\caption{\textbf{Video Evidence Indexing Prompt.}
The normal indexer receives the Video Preview, question, and answer choices.
During self-distillation, the teacher additionally receives privileged
information, including the question type and correct answer.}
\label{fig:selection_prompt}
\end{figure*}

\section{Additional Notes on Cost Estimates}
\label{sec:appendix_cost}

\begin{table}[h]
\centering
\resizebox{0.63\columnwidth}{!}{
\begin{tabular}{l c c c}
\toprule
\multirow{2}{*}{Method}
& \multirow{2}{*}{\makecell{Video Processed\\$(\#\mathrm{Frames} / \mathrm{Token})$}}
& \multirow{2}{*}{\makecell{Average Accuracy\\All Benchmarks}}
& \multirow{2}{*}{\makecell{Estimated Cost\\All Benchmarks}} \\
& & & \\
\midrule
Qwen3-VL-8B
& $2048_\mathrm{Max}/640_{\mathrm{Tok}}$
& 65.4
& $\sim\$87.8$ \\

Self-Distilled VEI
& $1\mathrm{fps}/24_{\mathrm{Tok}} + 32/128_{\mathrm{Tok}}$
& 66.9
& $\sim\$19.1$ \\

\bottomrule
\end{tabular}}
\caption{\textbf{Long-video accuracy and evaluation cost.}
The estimate covers 4,751 test questions across the three datasets.}
\label{tab:cost_long_video_comparison}
\vspace{-10pt}
\end{table}

The cost estimates use measured token counts and a fixed effective token rate. They are approximate and depend on encoding, cache reuse, and serving assumptions.

The comparison in Table~\ref{tab:cost_long_video_comparison} uses the same 8B backbone family and token-rate assumption for both rows. Self-Distilled VEI uses a \(50\mathrm{K}\) context limit, compared with \(256\mathrm{K}\) for the full-context baseline. The dollar values are approximate token-based estimates, rather than costs derived directly from these configured limits.

\section{Additional Discussion: Evidence Indexing}
\label{sec:appendix_evidence_indexing}

Long-video Video Evidence Indexing differs from standard keyframe sampling. In many QA examples, evidence may be spread across multiple moments: an object appears early, an action happens later, and the answer depends on comparing them. VEI therefore treats the preview-based indexing phase as Evidence Set construction, not isolated keyframe selection. The target list helps distribute the high-resolution frame budget across different evidence needs, while privileged self-distillation helps train this missing policy skill from student-generated rollouts.

\section{Qualitative Evidence and Failure Cases}
\label{sec:appendix_failure_cases}

\begin{figure}[h]
    \centering
    \includegraphics[width=1\linewidth]{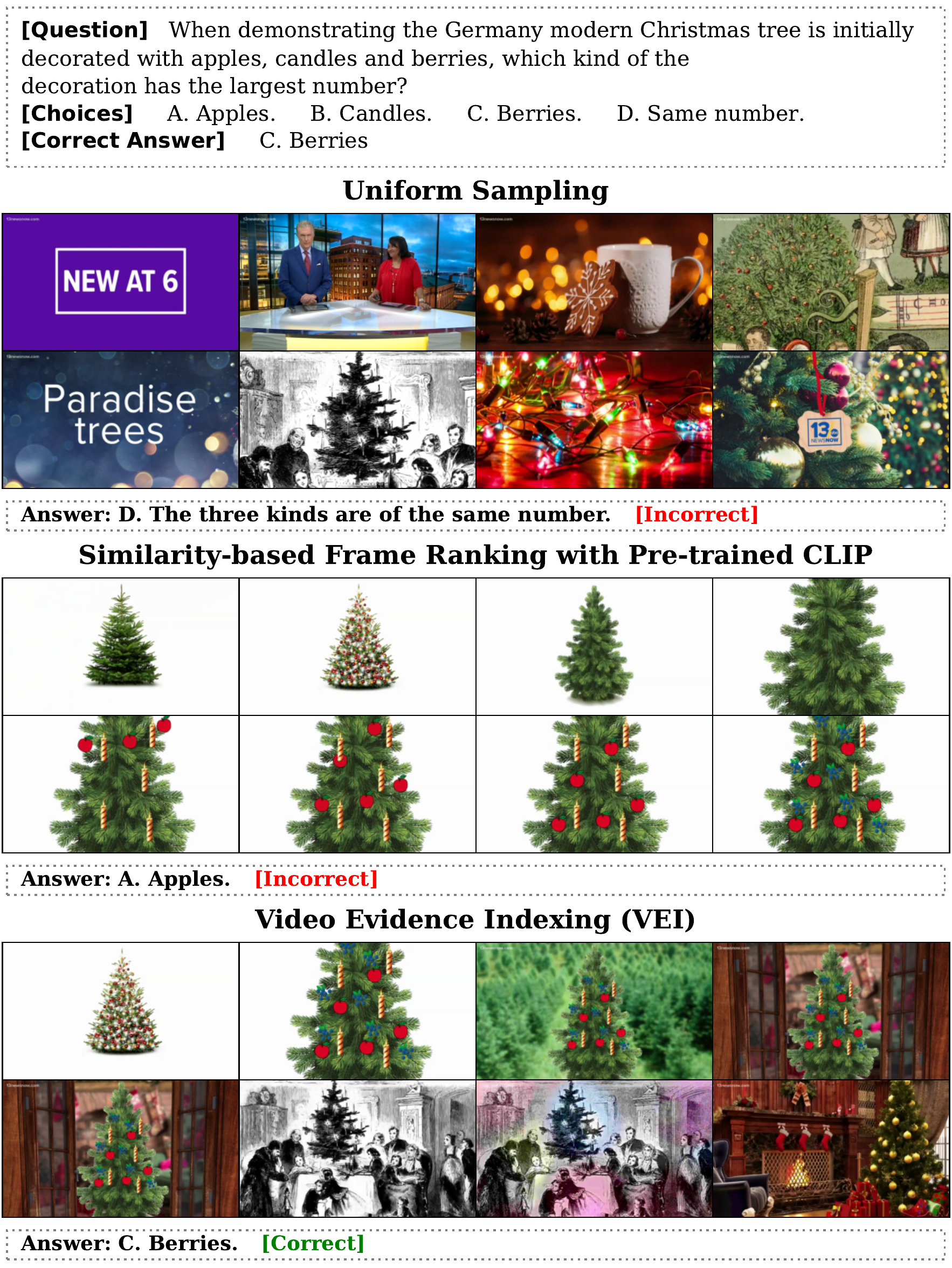}
    \vspace{-10pt}
    \caption{\textbf{EvSet Case Analysis.}
         Uniform sampling misses the close-up needed to compare decorations. CLIP retrieves repetitive tree views, while VEI includes detailed views of the completed decoration and answers correctly.}
    \label{fig:uniform_clip_VEI}
\end{figure}

In the example in Figure~\ref{fig:uniform_clip_VEI}, CLIP retrieves multiple similar views of the animated tree, whereas VEI covers more stages of the decoration sequence. This illustrates how target-wise selection can gather complementary observations for this question.

\begin{table}[h]
\centering
\small
\begin{tabular}{clp{0.63\textwidth}}
\toprule
Question & Outcome & Diagnosis \\
\midrule
61 & Hit, wrong & The selected evidence reaches the annotated event but misses its later outcome. \\
65 & Hit, wrong & The interaction is localized, but the answer misses the semantic detail that no bribe occurred. \\
73 & Miss, wrong & Selection concentrates on unrelated moments and misses the annotated event. \\
78 & Miss, correct & The answer is correct despite missing the annotated window; the preview or alternative evidence may provide support. \\
\bottomrule
\end{tabular}
\caption{\textbf{Representative LVBench cases.} These cases distinguish localization, incomplete event coverage, and downstream reasoning, and show that a single annotated window need not exhaust all useful evidence.}
\label{tab:failure_cases}
\end{table}

\section{Additional Evidence Set Ablation}
\label{sec:appendix_evset_ablation}

Table~\ref{tab:evidence_set_ablation} compares Training-Free VEI with a variant that removes the selected Evidence Set under the same configured context limits. With the preview fixed, this ablation measures the benefit of adding the selected high-resolution evidence.

The results show that the Evidence Set contributes consistently across all budgets. Removing it reduces the average score by \(11.1\), \(8.9\), \(5.2\), and \(5.0\) points under \(15\mathrm{K}\), \(20\mathrm{K}\), \(30\mathrm{K}\), and \(50\mathrm{K}\) context limits, respectively. The drop is largest in the lowest-budget setting, where the preview is useful for coarse temporal coverage but lacks enough visual detail for final answer verification. As the context budget increases, the preview itself becomes more informative, so the relative dependence on the Evidence Set decreases. However, the performance gap remains clear even at \(50\mathrm{K}\), showing that VEI is not only a low-resolution preview method. Its benefit comes from the combination of dense cheap coverage and targeted high-resolution evidence reinjection.

\begin{table*}[t]
\centering
\resizebox{\textwidth}{!}{
\begin{tabular}{l c cccc cc c}
\toprule
\multirow{2}{*}{Method} 
& \multirow{2}{*}{\makecell{Context\\Limit}}
& \multicolumn{4}{c}{VideoMME {\scriptsize w/o sub.}} 
& LVBench
& MLVU{\tiny MCQ}
& \multirow{2}{*}{\makecell{Average\\Score ($\%$)}} \\
\cline{3-8}
& 
& \scriptsize Short 
& \scriptsize Medium 
& \scriptsize Long 
& \scriptsize Overall 
& \scriptsize Train
& \scriptsize Test
& \\
\midrule

\rowcolor{gray!13}
\multicolumn{9}{c}{\textit{Use 5.9\% Context Width Across All Benchmarks}} \\
Training-Free VEI
& \multirow{2}{*}{$15\mathrm{K}$}
& $70.4$ & $53.9$ & $51.4$ & $58.6$ & $36.7$ & $46.2$
& $47.2$ \\
w/o Evidence Set
&
& $45.1$ & $43.1$ & $45.0$ & $44.4$ & $30.9$ & $32.9$
& $36.1$ {\color{blue}{$(-11.1)$}} \\
\midrule

\rowcolor{gray!13}
\multicolumn{9}{c}{\textit{Use 7.8\% Context Width Across All Benchmarks}} \\
Training-Free VEI
& \multirow{2}{*}{$20\mathrm{K}$}
& $73.2$ & $64.1$ & $55.9$ & $64.4$ & $46.1$ & $61.2$
& $57.2$ \\
w/o Evidence Set
&
& $59.9$ & $55.9$ & $55.7$ & $57.1$ & $40.7$ & $47.0$
& $48.3$ {\color{blue}{$(-8.9)$}} \\
\midrule

\rowcolor{gray!13}
\multicolumn{9}{c}{\textit{Use 11.7\% Context Width Across All Benchmarks}} \\
Training-Free VEI
& \multirow{2}{*}{$30\mathrm{K}$}
& $75.6$ & $67.2$ & $59.0$ & $67.3$ & $51.6$ & $63.5$
& $60.8$ \\
w/o Evidence Set
&
& $68.1$ & $62.4$ & $59.6$ & $63.4$ & $47.4$ & $56.0$
& $55.6$ {\color{blue}{$(-5.2)$}} \\
\midrule

\rowcolor{gray!13}
\multicolumn{9}{c}{\textit{Use 19.5\% Context Width Across All Benchmarks}} \\
Training-Free VEI
& \multirow{2}{*}{$50\mathrm{K}$}
& $77.7$ & $70.3$ & $63.6$ & $70.5$ & $53.5$ & $66.7$
& $63.6$ \\
w/o Evidence Set
&
& $72.0$ & $66.1$ & $60.2$ & $66.1$ & $50.8$ & $59.0$
& $58.6$ {\color{blue}{$(-5.0)$}} \\
\bottomrule
\end{tabular}}
\caption{\textbf{Effect of Evidence Set access under matched configured context limits.}
For each low-resolution preview budget, we compare training-free VEI with the same model after removing access to the selected high-resolution Evidence Set. Blue values report the absolute average-score drop caused by removing Evidence Set frames.}
\label{tab:evidence_set_ablation}
\end{table*}

\begin{table*}[t]
\centering
\caption{Training Configuration for Privileged Self-Distillation}
\label{tab:sd_training_config}
\begin{tabular}{l c}
\toprule
\textbf{Parameter} & \textbf{SD} \\
\midrule
Learning Rate & $1 \times 10^{-4}$ \\
Effective Batch Size & $8$ \\
\midrule
LoRA Rank ($r$) & $64$ \\
LoRA Alpha ($\alpha$) & $128$ \\
LoRA Target Modules & 
\makecell{q\_proj, k\_proj, v\_proj, o\_proj,\\
gate\_proj, up\_proj, down\_proj} \\
\midrule
Max Completion Length & $1024$ \\
\midrule
Number of Generations per Prompt & $1$ \\
Sampling Temperature & $1.0$ \\
KL Coefficient ($\beta$) & -- \\
Training Steps & $1500$ \\
\bottomrule
\end{tabular}
\end{table*}

\clearpage
\section{Accuracy by Question Type}
\label{sec:appendix_question_types}

Tables~\ref{tab:question_types_videomme}--\ref{tab:question_types_mlvu} compare Uniform FS, Training-Free VEI, and Self-Distilled VEI across each dataset's question types and all four preview resolutions. Each comparison uses the same preview and 32-frame evidence budget. Bold marks the highest displayed accuracy within each question type and preview resolution, including ties. Overall results appear in Table~\ref{tab:main_result}.

\paragraph{VideoMME.}
At one token per frame, self-distillation improves over Training-Free VEI on 10 of 12 question types, with the largest gain in temporal perception. At 24 tokens, gains remain clear in action reasoning and OCR, while counting and several perception categories favor Training-Free VEI. Training therefore changes the capability profile rather than improving every question type equally.

\noindent\begin{minipage}{\textwidth}
\centering
\small
\setlength{\tabcolsep}{2pt}
\renewcommand{\arraystretch}{1.08}
\begin{tabular*}{\linewidth}{@{\extracolsep{\fill}}l*{12}{r}@{}}
\toprule
\multirow{2}{*}{Question type} & \multicolumn{3}{c}{1 token/frame} & \multicolumn{3}{c}{6 tokens/frame} & \multicolumn{3}{c}{12 tokens/frame} & \multicolumn{3}{c}{24 tokens/frame} \\
\cmidrule(lr){2-4}\cmidrule(lr){5-7}\cmidrule(lr){8-10}\cmidrule(lr){11-13}
 & U & TF & SD & U & TF & SD & U & TF & SD & U & TF & SD \\
\midrule
Action Reasoning & 56.1 & 50.5 & \textbf{61.1} & 57.9 & 55.8 & \textbf{62.8} & 58.3 & 58.6 & \textbf{64.6} & 60.0 & 61.8 & \textbf{67.7} \\
Action Recognition & 57.8 & 56.2 & \textbf{63.3} & 62.0 & 63.9 & \textbf{65.5} & 65.2 & 64.9 & \textbf{66.1} & 65.2 & 67.1 & \textbf{69.5} \\
Attribute Perception & 77.9 & 70.3 & \textbf{79.3} & \textbf{81.1} & 75.2 & 80.6 & 82.0 & 79.3 & \textbf{82.4} & 78.8 & \textbf{83.3} & 82.4 \\
Counting Problem & 36.2 & \textbf{39.2} & 38.1 & 39.9 & \textbf{43.7} & 41.4 & 42.9 & \textbf{48.1} & 44.4 & 41.4 & \textbf{48.1} & 42.5 \\
Information Synopsis & \textbf{82.0} & 76.5 & 79.9 & \textbf{82.0} & 77.7 & 81.1 & \textbf{83.9} & 82.4 & 82.0 & 84.8 & 83.9 & \textbf{85.9} \\
Object Reasoning & \textbf{61.2} & 58.2 & \textbf{61.2} & \textbf{64.5} & 63.4 & 63.2 & 64.8 & 65.2 & \textbf{66.5} & 66.7 & \textbf{68.9} & 67.8 \\
Object Recognition & \textbf{67.0} & 64.1 & \textbf{67.0} & 68.6 & \textbf{70.9} & 69.8 & 69.8 & 70.9 & \textbf{71.2} & 70.6 & 75.7 & \textbf{75.8} \\
OCR Problems & \textbf{71.9} & 61.2 & 68.4 & 71.9 & 70.5 & \textbf{74.1} & 72.7 & 73.4 & \textbf{75.5} & 72.7 & 79.1 & \textbf{83.3} \\
Spatial Perception & \textbf{66.7} & 59.3 & 63.0 & \textbf{70.4} & 59.3 & 64.8 & \textbf{70.4} & 68.5 & 66.7 & 64.8 & \textbf{66.7} & 65.8 \\
Spatial Reasoning & \textbf{82.1} & 75.0 & 75.0 & \textbf{85.7} & 78.6 & 80.4 & \textbf{83.9} & 78.6 & 75.0 & 80.4 & \textbf{82.1} & 81.6 \\
Temporal Perception & 65.5 & 50.9 & \textbf{72.7} & 76.4 & 70.9 & \textbf{78.2} & 76.4 & \textbf{78.2} & 74.6 & 78.2 & \textbf{81.8} & 79.4 \\
Temporal Reasoning & 48.6 & 42.9 & \textbf{52.0} & 50.9 & 52.5 & \textbf{53.7} & 57.1 & 57.6 & \textbf{61.0} & 59.9 & 65.0 & \textbf{67.1} \\
\bottomrule
\end{tabular*}
\captionof{table}{\textbf{Question-type accuracy on VideoMME.} Scores (\%) compare Uniform FS (U), Training-Free VEI (TF), and Self-Distilled VEI (SD) at each preview resolution.}
\label{tab:question_types_videomme}
\end{minipage}
\par\vspace{8pt}

\paragraph{LVBench.}
Training-Free VEI exceeds uniform sampling on temporal grounding at every preview resolution. Self-distillation further improves this category across all four resolutions and, at 24 tokens, improves all six question types over Training-Free VEI. The gains extend beyond temporal grounding to information retrieval and reasoning.

\noindent\begin{minipage}{\textwidth}
\centering
\small
\setlength{\tabcolsep}{2pt}
\renewcommand{\arraystretch}{1.08}
\begin{tabular*}{\linewidth}{@{\extracolsep{\fill}}l*{12}{r}@{}}
\toprule
\multirow{2}{*}{Question type} & \multicolumn{3}{c}{1 token/frame} & \multicolumn{3}{c}{6 tokens/frame} & \multicolumn{3}{c}{12 tokens/frame} & \multicolumn{3}{c}{24 tokens/frame} \\
\cmidrule(lr){2-4}\cmidrule(lr){5-7}\cmidrule(lr){8-10}\cmidrule(lr){11-13}
 & U & TF & SD & U & TF & SD & U & TF & SD & U & TF & SD \\
\midrule
entity recognition & 42.7 & 35.3 & \textbf{42.8} & 44.0 & 47.0 & \textbf{48.6} & 48.9 & 52.6 & \textbf{54.1} & 53.8 & 55.4 & \textbf{57.8} \\
event understanding & 38.2 & 35.1 & \textbf{45.0} & 39.9 & 45.0 & \textbf{49.3} & 42.2 & 48.2 & \textbf{50.5} & 47.1 & 49.9 & \textbf{55.0} \\
key information retrieval & 39.5 & 37.5 & \textbf{47.4} & 45.0 & 46.7 & \textbf{49.8} & 48.8 & 54.6 & \textbf{55.7} & 55.0 & 55.3 & \textbf{62.1} \\
reasoning & 35.8 & 34.3 & \textbf{42.8} & 36.8 & 36.8 & \textbf{51.2} & 42.8 & 46.3 & \textbf{50.8} & 44.8 & 44.8 & \textbf{52.7} \\
summarization & 31.0 & \textbf{43.1} & 32.8 & 34.5 & \textbf{46.6} & 44.8 & 39.7 & \textbf{46.6} & \textbf{46.6} & 39.7 & 44.8 & \textbf{46.5} \\
temporal grounding & 37.3 & 56.4 & \textbf{60.0} & 40.9 & 51.8 & \textbf{58.2} & 45.9 & 56.4 & \textbf{59.6} & 48.6 & 56.4 & \textbf{59.9} \\
\bottomrule
\end{tabular*}
\captionof{table}{\textbf{Question-type accuracy on LVBench.} Scores (\%) compare Uniform FS (U), Training-Free VEI (TF), and Self-Distilled VEI (SD) at each preview resolution.}
\label{tab:question_types_lvbench}
\end{minipage}
\par\vspace{8pt}

\clearpage
\paragraph{MLVU.}
At one token per frame, self-distillation improves over Training-Free VEI across all nine question types. At 24 tokens, the largest gains occur in \textit{ego} and \textit{count}, while gains are smaller or mixed in the other categories. Together, the three datasets show that training helps across varied question types under extreme preview compression, with more task-dependent gains at higher resolution.

\noindent\begin{minipage}{\textwidth}
\centering
\small
\setlength{\tabcolsep}{2pt}
\renewcommand{\arraystretch}{1.08}
\begin{tabular*}{\linewidth}{@{\extracolsep{\fill}}l*{12}{r}@{}}
\toprule
\multirow{2}{*}{Question type} & \multicolumn{3}{c}{1 token/frame} & \multicolumn{3}{c}{6 tokens/frame} & \multicolumn{3}{c}{12 tokens/frame} & \multicolumn{3}{c}{24 tokens/frame} \\
\cmidrule(lr){2-4}\cmidrule(lr){5-7}\cmidrule(lr){8-10}\cmidrule(lr){11-13}
 & U & TF & SD & U & TF & SD & U & TF & SD & U & TF & SD \\
\midrule
anomaly\_reco & 48.7 & 46.2 & \textbf{51.3} & 53.9 & 48.7 & \textbf{56.4} & 53.9 & \textbf{56.4} & \textbf{56.4} & 51.3 & 53.9 & \textbf{56.4} \\
count & 26.7 & 11.7 & \textbf{30.0} & 30.0 & 21.7 & \textbf{53.3} & 31.7 & 25.0 & \textbf{56.7} & 30.0 & 35.0 & \textbf{55.0} \\
ego & 56.6 & 41.5 & \textbf{60.4} & 54.7 & 58.5 & \textbf{77.4} & 58.5 & 58.5 & \textbf{79.3} & 60.4 & 56.6 & \textbf{79.3} \\
needleQA & 50.0 & 50.0 & \textbf{60.0} & 58.3 & \textbf{80.0} & 78.3 & 61.7 & 81.7 & \textbf{83.3} & 71.7 & \textbf{85.0} & 83.3 \\
order & 42.9 & 30.0 & \textbf{62.9} & 55.7 & 65.7 & \textbf{77.1} & 61.4 & 68.6 & \textbf{84.3} & 60.0 & 75.7 & \textbf{82.9} \\
plotQA & 42.0 & 50.0 & \textbf{56.0} & 48.0 & 50.0 & \textbf{66.0} & 52.0 & \textbf{62.0} & \textbf{62.0} & 60.0 & 68.0 & \textbf{70.0} \\
sportsQA & 41.7 & 41.7 & \textbf{47.2} & 41.7 & \textbf{50.0} & 38.9 & 47.2 & \textbf{50.0} & 47.2 & 44.4 & \textbf{50.0} & 47.2 \\
topic\_reasoning & \textbf{85.7} & 79.1 & 81.3 & \textbf{84.6} & \textbf{84.6} & 80.2 & 84.6 & \textbf{86.8} & 84.6 & 83.5 & \textbf{85.7} & 84.6 \\
tutorialQA & 53.5 & 51.2 & \textbf{60.5} & 44.2 & \textbf{69.8} & 65.1 & 58.1 & 60.5 & \textbf{74.4} & 58.1 & \textbf{67.4} & \textbf{67.4} \\
\bottomrule
\end{tabular*}
\captionof{table}{\textbf{Question-type accuracy on MLVU.} Scores (\%) compare Uniform FS (U), Training-Free VEI (TF), and Self-Distilled VEI (SD) at each preview resolution.}
\label{tab:question_types_mlvu}
\end{minipage}
\par\vspace{8pt}

\section{Limitations}
\label{sec:limitations_ethics}

Our evaluation focuses on a small set of 8B VLMs, and results may vary with model scale and pretraining. We report single-run evaluations without estimates of training-seed variability. Broader adaptive-selector comparisons would extend our uniform, CLIP-retrieval, and nested-VEI controls.

Our CLIP$^{\dagger}$ baseline uses a publicly pretrained CLIP model to embed the question and each candidate frame independently. We rank frames by question--image embedding similarity and select the top 32, retaining the same preview and high-resolution evidence budget as Training-Free VEI.

The localization analysis measures temporal evidence recovery; direct measurements of target coverage and visual redundancy remain outside this study. The distilled checkpoint performs both indexing and answering, so the QA results characterize the complete pipeline without separately quantifying changes in answering behavior.

Very low-resolution previews can obscure evidence needed for indexing, while a fixed-size Evidence Set may under-cover extended events. VEI processes completed videos and does not address causal online streaming. Our efficiency analysis focuses on context capacity; end-to-end latency and memory across both inference calls remain to be characterized.

\end{document}